\documentclass[a4paper,12pt]{article}

\usepackage[utf8]{inputenc} 
\usepackage{amsmath,amssymb} 
\usepackage{graphicx} 
\usepackage[numbers]{natbib} 
\usepackage[hidelinks]{hyperref} 
\usepackage{tikz} 
\usepackage{pgfplots} 
\pgfplotsset{compat=1.18} 

\title{Assessing AI Adoption and Digitalization in SMEs: A Framework for Implementation}
\author{
    Serena Proietti\\
    University of Rome Tor Vergata, E.N.I.A\\
    \texttt{serena.proietti@uniroma2.it}
    \and
    Roberto Magnani\\
    co-founder TOPForGrowth, E.N.I.A\\
    \texttt{robertomagnani@aol.com}
}
\date{}

\begin{document}

\maketitle

\begin{abstract}
The primary objective of this research is to examine the current state of digitalization and the integration of artificial intelligence (AI) within small and medium-sized enterprises (SMEs) in Italy. There is a significant gap between SMEs and large corporations in their use of AI, with SMEs facing numerous barriers to adoption. This study identifies critical drivers and obstacles to achieving intelligent transformation, proposing a framework model to address key challenges and provide actionable guidelines.
\end{abstract}


\section{Introduction}
In recent times, Artificial Intelligence (AI) has transformed numerous industries, providing unparalleled possibilities for enhancing processes and improving decision-making. While many large companies have started incorporating AI into their workflows, small and medium-sized enterprises (SMEs) have not adopted these technologies as widely, even though they could benefit substantially from improvements in efficiency and competitiveness. Specifically, SMEs face considerable obstacles, largely due to the smaller amounts of data they handle in comparison to larger businesses, as well as the generally lower quality of the data they have available.

It is important to underline that the definition of SMEs varies significantly across countries, particularly in terms of revenues, organizational size, and the number of employees. These distinctions are critical when discussing both the opportunities and limitations associated with AI implementation in SMEs. 
The European Union (EU) classifies SMEs into three categories: (i) micro enterprises (fewer than 10 employees and a turnover of less than 2 million euros), (ii) small enterprises (fewer than 50 employees and a turnover of less than 10 million euros), and (iii) medium-sized enterprises (fewer than 250 employees and a turnover of less than 50 million euros) \citep{munro2013sme}.
Since the research will refer to a group of Italian small and medium-sized enterprises, it will use the aforementioned reference. However, numerous countries and organizations outside the EU define this term in significantly different ways.

This research provides important insights for the following research questions:
\begin{itemize}
\item \textbf{RQ1}: What is the current landscape about the implementation of AI techniques and digitalization in SMEs using the sample of 36 organizations operating in 14 industries?

\item \textbf{RQ2}: What could be a possible framework for AI implementation in SMEs?

\end{itemize}
By addressing these research questions, the study provides the following main contributions:

\begin{enumerate}
\item Identify the uses of AI and the level of digitalization within small and medium-sized organizations proposed, undestanding barriers and opportunities.

\item  Identify guidelines for AI implementation in SMEs that takes into account the specific needs and limitations of the sector.

\end{enumerate}

By identifying the challenges and barriers that SMEs encounter in AI adoption, the study will equip managers, professionals, and decision-makers with targeted strategies to effectively overcome these obstacles. 
Additionally, the research will provide valuable insights into the management of AI implementation processes, including employee participation and training, thereby ensuring successful integration and maximizing AI's impact on organizational performance.

The rest of the document is organized as follows:
Section II provides a literature review about Artificial Intelligence and its application in SMEs. Section III describes the research methodology used to conduct the survey. Section IV explores the findings and  Section V describes the conceptual framework. At the end, section VI summarizes final discussion.

\section{Literature Review}
Over the years, extensive research has focused on applying Artificial Intelligence (AI) in small and medium-sized enterprises (SMEs) to bridge the informational gap between these organizations and larger corporations. Despite these efforts, significant challenges persist, and further investigation is needed to identify and address the primary obstacles SMEs face when adopting AI technologies. While existing literature reviews have offered valuable insights, there is a clear need for more comprehensive research to fully explore the complexities and barriers involved in AI integration within SMEs.

The journey of AI began in 1956, when John McCarthy first coined the term "Artificial Intelligence", defining it as the science and engineering of creating intelligent machines \citep{mccarthy2006proposal}. Not long after, in 1959, the introduction of the first industrial robot, Unimate, marked a significant step forward \citep{day2018robotics}. Unimate captured the interest of General Motors, which adopted it to automate tasks on the assembly line, illustrating the transformative potential of AI in industry.

Progress continued in 1964 with the development of the first chatbot at the MIT Artificial Intelligence Laboratory \citep{shum2018eliza}. This was soon followed by another milestone: between 1966 and 1972, Stanford Research Institute developed Shakey, the first robot to combine logical reasoning with physical action \citep{nilsson1984shakey}. These early successes laid the foundation for AI applications in industrial settings, though they also preceded a period known as the \textit{AI Winter}—a phase marked by diminished optimism and limited advancements due to the constraints of available data and early-stage technology.

These led to a decline in interest in the field that lasted for approximately 30 years. The resurgence of AI can be attributed to around 1996, with the debut of Deep Blue, a supercomputer designed by IBM.

Nowadays, with the advancements in Neural Networks (NN) and the vast amounts of available data, AI applications have increased exponentially.
The development of advanced NN architectures, including Recurrent Neural Networks (RNN), Long Short-Term Memory (LSTM), Convolutional Neural Networks (CNN), and Transformers with the self-Attention mechanisms\citep{vaswani2017attention}, has led to a substantial evolution in the use of AI \citep{cong2023review},\citep{bahdanau2014neural}. These innovations have enhanced the ability of AI systems to process complex data patterns, improve predictive accuracy, and manage sequential data more effectively. As a result, AI has expanded into a wide range of applications, from Natural Language Processing (NLP) \citep{nadkarni2011natural} and Image Recognition (IR) \citep{chauhan2018convolutional} to time series forecasting and beyond, significantly broadening its scope and potential applications.

In adding, major tech giants in the AI market are actively developing numerous cloud-based solutions to integrate artificial intelligence into businesses. Companies like Amazon Web Services (AWS), Microsoft Azure, Google Cloud, and IBM Cloud are leading this effort, offering a range of AI services designed to help businesses adopt AI technologies and facilitate digital transformation.

In addition, foundation models play a central role in these cloud-based solutions, enabling businesses to harness the power of large-scale AI without the need for extensive in-house resources. 

In line with this trend, fine-tuning serves as a critical mechanism for optimizing the performance of AI services provided by these cloud platforms, allowing businesses to customize pretrained foundation models to meet specific requirements by utilizing smaller, labeled datasets, thereby delivering more precise and contextually relevant solutions.

Despite the progress in this field, SMEs face many barriers in implementation.

Many studies identified different challenges over in the implementation of AI in SMEs.
According to the SLR citepd \citep{oldemeyer2024investigation}, the predominant issue identified across the selected papers is knowledge. In fact, this obstacle is mentioned and thus much more often than all other. The second most common reason for the hesitation to implement AI in SMEs are the costs.
The maturity level within their own organization stands out as the third most frequently referenced barrier  for SMEs when embarking on AI adoption.
Furthermore, if potential users view the implementation of a new technology as difficult and complex, they are less inclined to adopt it. Numerous studies have demonstrated that the ease of use of a new technology plays a crucial role in its acceptance. At the same time the advantages of adopting a new technology motivate companies to adopt it. \citep{9908467}. 

Other aspects regard top management's commitment, organizational readiness, external support, employee adoption, and competitive pressure as key determinants that influence the adoption of AI technologies in small and medium enterprises. A clear understanding of the key drivers for AI implementation can help managers and owners of SMEs to take appropriate actions and initiatives to adopt AI technologies \citep{ingalagi2021artificial}.

Nevertheless many studies indicate the need for more research to unlock AI's potential in this sector, for exploring SME characteristics and drivers. The research should focus on simplifying AI solutions to make them more accessible, as current research shows SMEs struggle with the complexity of AI despite its benefits \citep{hansen2021artificial}.

\section{Methodology}
\begin{figure}[h!]
    \centering
    \includegraphics[width=0.7\textwidth]{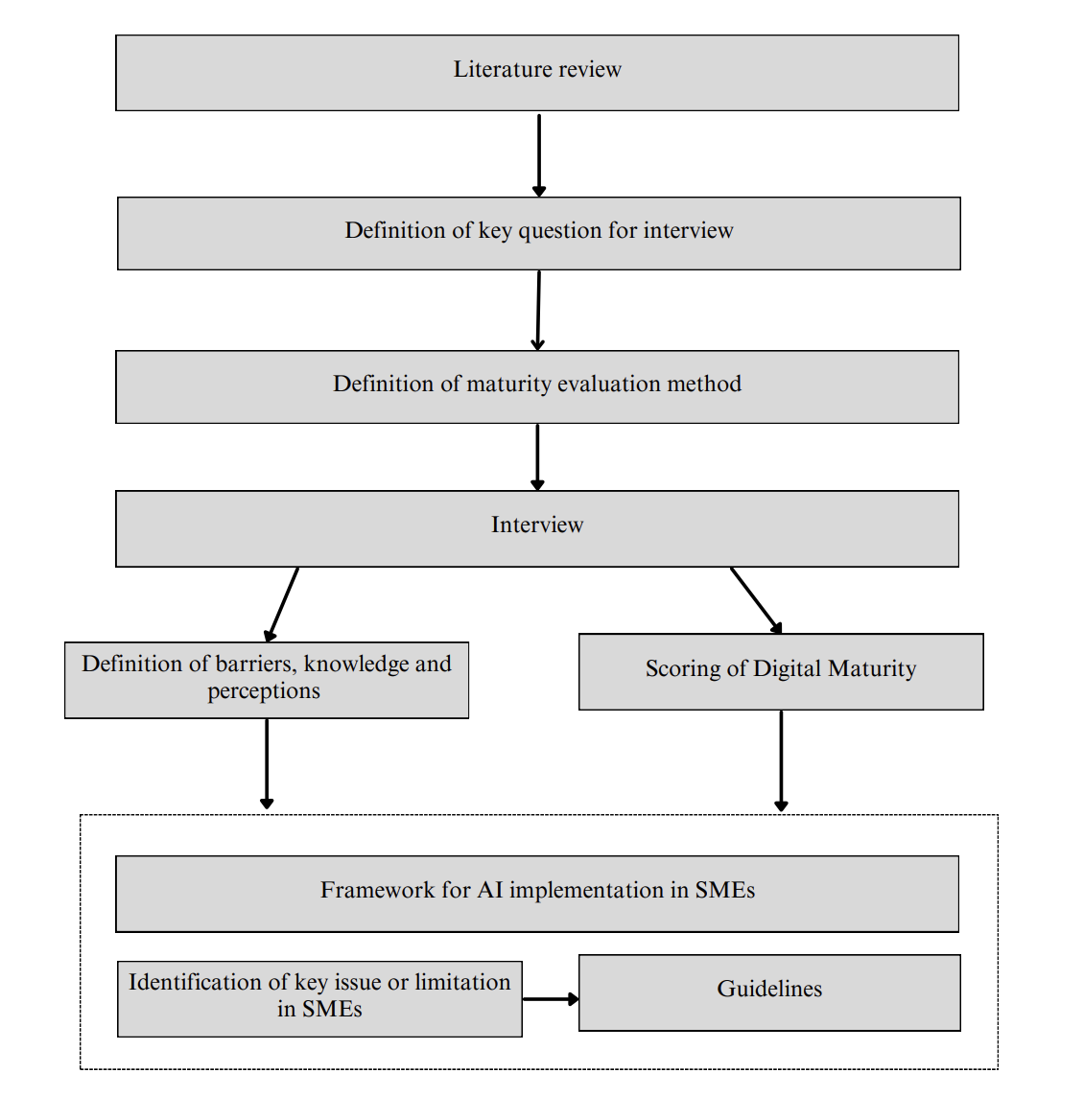}
    \caption{Research method} 
    \label{method} 
\end{figure}

The research methodology presented in Figure \ref{method} was followed in this study. The proposed model was developed by reviewing the state of the art and defining key questions for interviews. Subsequently, a method for maturity evaluation was defined based on these questions.

The interview phase involved managers from SMEs across different sectors. These activities allowed us to identify existing barriers, knowledge, and perceptions about AI. Based on these insights, the framework was defined, identifying key gaps and providing guidelines for SMEs.

\subsection{Sample design}
To determine the digital transformation status and AI usage within a group of small and medium-sized enterprises (SMEs) in Italy, a project was launched in collaboration with non-profit organization in Italy that represents and supports the interests of small and medium enterprises. A survey was submitted to a group of 36 SMEs from various sectors such as Commerce, Communication and Services, Construction, Installations (Thermal-electric-TV), Information Technology, Metalworking, and others.

In each SME, interviews were conducted through a questionnaire that included an initial set of closed-ended questions related to the organization's description, in order to ensure proper segmentation during the analysis phase.

\subsection{Data Collection and analysis}
The survey consists of a total of 15 questions, primarily closed-ended questions with the option to provide a qualitative response when none of the alternatives are selected. Most of the companies in the sample are small in size, with nearly all reporting a turnover of less than 3 million. There is a high diversity among the sectors represented, with the predominant sector being commerce. 

The survey questions can be divided into four main areas: organization description, level of digitalizationy, barriers in the implementation and knowledge of AI, possible positive and negative impacts and perception regarding the importance of Artificial Intelligence. 

Some questions can directly influence scores, while some others provide insight into maturity but indirectily influence scores as shown in table \ref{table1}.

The results were analyzed using Excel, with segmentation based on size, revenue, and number of employees.

Based on the responses, a digital maturity level was assigned from 1 to 4:
\begin{itemize}
\item Very Low Maturity (Minimal digital tools, no AI)
\item Low Maturity (Basic digital tools, limited knowledge and understanding of AI)
\item Medium Maturity (Some advanced tools, moderate use of AI for business processes)
\item High Maturity (Advanced digital tools, strategic use of AI)

\end{itemize}

Each question was assigned a coefficient related to the impact of that specific question on the level of digital maturity. Subsequently, each possible response provided had a normalized score related to the weight of the response on the level of digital maturity.

\begin{table}[hb]
\begin{center}
\caption{Survey Structure Analysis}\label{table1}
\begin{tabular}{ccc}
\textbf{Survey Areas}	& \textbf{Questions}	& \textbf{scores' Influence}\\ \hline
\
Organization Description	& 3		& Indirect\\ \hline
Level of Digitalizations\textsuperscript{1} 	& 5		& Direct\\ \hline

Barriers and Knowledge & 	3	& Direct\\ \hline
Impact and Perception of AI 		& 4			& Indirect\\
\end{tabular}
\end{center}
\end{table}
\section{Findings}

The level of digitalization among SMEs has been assessed by evaluating the use of less advanced tools such as websites, information systems, and Excel, for instance, up to the utilization of more advanced technologies such as AI-based systems or digital twins.

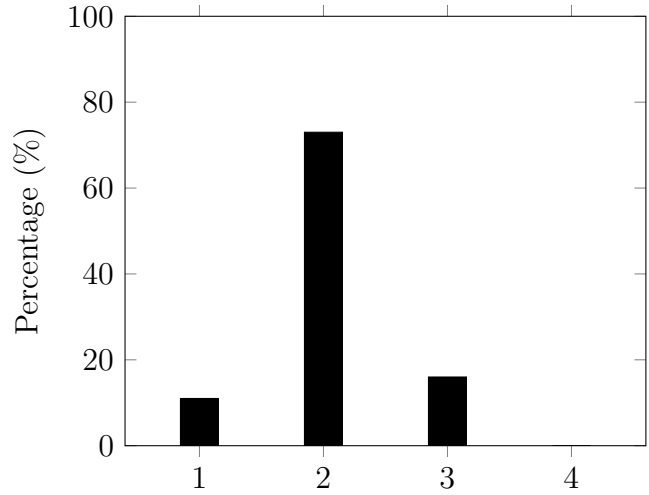
\begin{figure} 
    \centering
    \begin{tikzpicture}
        \begin{axis}[
            ybar,
            bar width=0.5cm,
            symbolic x coords={1, 2, 3, 4},
            xtick=data,
            ymin=0,
            ymax=100,
            ylabel={Percentage (\%)},
            every axis plot/.append style={fill=black},
            enlarge x limits=0.2
        ]
            \addplot[fill=black] coordinates {(1,11) (2,73) (3,16) (4,0)};
        \end{axis}
    \end{tikzpicture}
    \caption{Distribution of SMEs Across Categories} 
    \label{grafico}
\end{figure}

The results are shown in figure \ref{grafico}. As can be observed, none of the small and medium-sized enterprises was classified into the category of high digital maturity, and most were placed in the category with limited knowledge of AI without utilizing it for business processes.

The use of Excel is fairly widespread for various functionalities, with only 16.7\% of respondents indicating that they do not utilize it. Slightly less than 14\% of the enterprises do not have a website. A website provides significant online presence for a business; without it, the company may not be easily found or visible to potential customers on the internet. This could diminish the opportunity to reach a broader audience and leverage potential business opportunities.

Only half of the respondents actively use social media on a daily basis. Social media has become a crucial communication and interaction channel for businesses. A lack of activity on social media can result in the company lacking visibility and a significant online presence, limiting its ability to engage with a broader audience and interact with current and potential customers on a popular platform widely used by many.

A small percentage of companies currently utilize advanced technologies to support their processes, such as IoT, robotics, augmented reality, and virtual reality, while 2.8\% reported not using any of the mentioned technologies. 

Only 13.9\% leverage AI to support their processes. Furthermore, nearly half of the respondents have not made, nor are they planning to make, any technological investments.

Without adequate investments, a company may become less competitive compared to its rivals that adopt advanced technological solutions. This could restrict the company's ability to attract customers, increase sales, and maintain a prominent market position.

Innovation is a key element for business success. Technological investments enable companies to keep pace with the latest industry trends, develop new products or services, and adopt cutting-edge business practices.

In the absence of technological investments, a company risks falling behind its competitors and missing out on innovation opportunities that could facilitate significant growth.

Only 11\% of respondents are familiar with Generative AI and use it regularly.

Generative AI can significantly transform work by automating tasks that currently occupy 60-70\% of employees' time, this acceleration is driven by AI's improved ability to understand natural language, crucial for tasks making up 25\% of work time. As a result, generative AI impacts knowledge-based, higher-wage jobs more than other occupations.
\citep{chui2023economic}

The majority of respondents believe that Generative AI can primarily support customer communication and assistance, as well as content creation for social media or marketing.

Conducting activities in a traditional manner, without leveraging the opportunities offered by advanced technologies to automate processes and improve operational efficiency, may lead to increased reliance on manual processes that require more time and resources, thereby increasing costs and reducing overall productivity. More than 70\% of companies have never utilized Generative AI to support their processes.

However, the lack of utilization of Generative AI by companies can also be viewed as an opportunity. Companies that begin to explore the adoption of this technology could benefit from being at the forefront of their industry and innovating their business processes. It would be beneficial to provide companies with information and concrete evidence regarding the benefits and opportunities of Generative AI, also offering support and specialized consulting to guide them in implementing solutions tailored to their needs.

More than half of the respondents stated that they do not have a clear understanding of how AI can be used to support their processes.

To overcome these challenges and develop a clearer idea of how artificial intelligence can be utilized in business processes, it is advisable to explore available resources, such as case studies, articles, webinars, and conferences on the subject. Collaborating with AI experts or specialized consultants could also help identify specific opportunities and tailor solutions to the business context.

Integrating AI solutions into business processes may require careful analysis, the identification of specific opportunities and challenges, and the implementation of appropriate models and algorithms. The complexity of such integration can create uncertainty or difficulty in understanding potential applications.

The majority of respondents believe that the potential positive effects include cost reduction and an improvement in the company's competitiveness. As for the negative effects, the loss of humanity is among the most feared impacts. This highlights the importance of developing increasingly human-centered AI systems. The fact that AI is revolutionizing the way we work is evident, making it crucial to understand the added value that humans bring in the use of these intelligent systems.

In fact, just over 22\% of respondents fear a reduction in revenue or a decrease in employment.

These concerns stem primarily from the potential to automate processes, replace human skills, and uncertainties about the economic impact and the job market.

However, it is important to emphasize once again that the introduction of AI can enhance human resources in new roles, offering the opportunity to expand the company's business.

Among the barriers feared by companies when introducing GAI-based solutions, as revealed by the questionnaire, the majority of respondents are concerned about excessively high costs and lack of expertise. 

Companies, in fact, without a clear understanding of the potential benefits of adopting these new technologies, may worry that the initial costs will be too high compared to the expected improvements. 

Additionally, the lack of knowledge could lead to a misperception of the costs and risks associated with implementing AI-based solutions, further contributing to the apprehension of businesses.

\section{Framework}
For defining the framework, a bottom-up approach was employed, starting with all aspects that should be considered in the development of an AI-based project within an SME. Additionally, the approach drew upon a study conducted in Singapore in collaboration with the World Economic Forum Centre for the Fourth Industrial Revolution, the Infocomm Media Development Authority (IMDA) and the Personal Data Protection Commission (PDPC) was teakes as reference. \citep{isago}.

The guiding principles in the definition of Key factor and framework structure: 
\begin{itemize}
    \item AI system should underline the relavance of human aspect 
    \item AI system should be relevant for the organization
    \item AI system should be technically feasible 
\end{itemize}

\subsection{Key factors}
In the current landscape, the integration of AI in SMEs represents a substantial challenge, as these entities faces many obstacles, including limited resources availability, limited knowledge of the topic and often resistance to change. The framework's development initially focused on identifying the distinctions between SMEs and large corporations, with the objective of designing a tailored tool.

\begin{table}[hb]
\begin{center}
\caption{Differences between SMEs and Big Companies}
\label{differences}
\begin{tabular}{cccc}
\textbf{SMEs}	& \textbf{Big Companies}	\\ \hline
Limited budget	& More extensive financial resources \\ \hline
Limited Knowledge & Dedicated teams for AI projects \\ \hline
Cultural resistance & More innovation-oriented culture \\ \hline
Limited tech infrastructure & Easier Integration \\ \hline
\end{tabular}
\end{center}
\end{table}

As shown in table \ref{differences}, there are several challenges that an SME, unlike a large company, must overcome in integration of these systems. A notable aspect involves culture; infact, technical knowledge is not enough without an emotional intelligence and an ability to different cultural contexts. In fact, this is crucial because AI adoption, often leads to significant changes in working processes, which can bring anxiety, uncertainty, and resistance. 

In adding, the Italian environment, particularly in sectors like SMEs and social organizations, often shows resistance to significant technological changes that might disrupt traditional work cultures. However, this resistance is counterbalanced by a practical recognition that technology, especially AI, can bring benefits when implemented with cultural sensitivity. \citep{magnani2024}.

\begin{figure}[h!]
    \centering
    \includegraphics[width=0.7\textwidth]{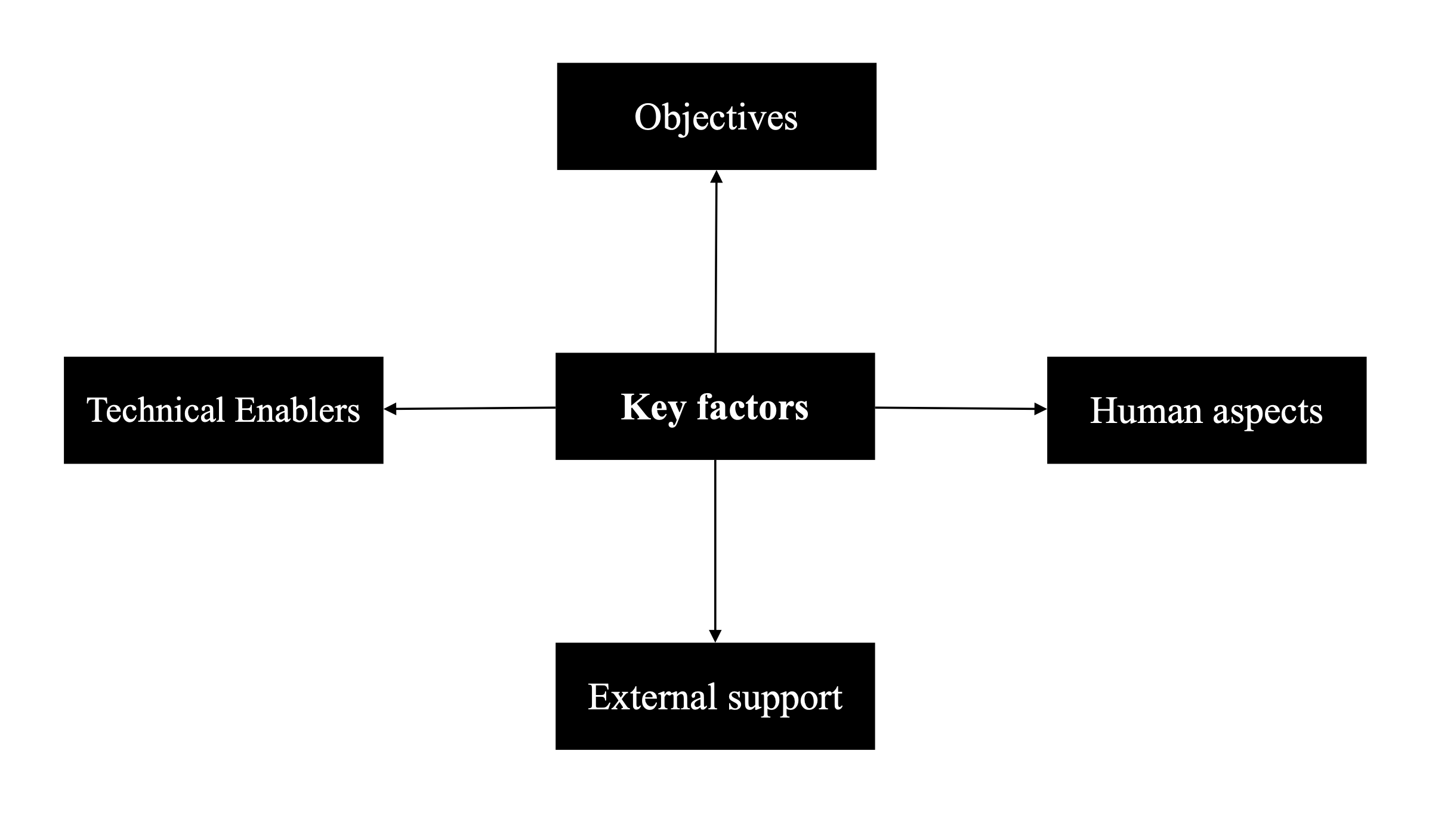}
    \caption{Key factors to evaluate for new AI implementation} 
    \label{key factors} 
\end{figure}

The framework includes several key factors that are crucial to identify and understand (figure \ref{key factors}). These factors are necessary for developing targeted strategies and support programs that facilitate successful AI implementation and integration.

\subsubsection{Objectives}
accurate objective-setting is essential for a successful project; unclear or unrealistic goals can easily lead to project failure. While framing the problem question is a key step, it's equally critical to establish the problem's priority, as this helps maintain focus and commitment throughout the project. One approach to ensuring clarity in goal definition is to start by asking what problem are we aiming to solve, this question encourages a deeper exploration, using the SMART approach to craft goals that are Specific, Measurable, Achievable, Relevant, and Time-bound. Well-defined objectives minimize misunderstandings, guide effective decision-making, and significantly increase the project's likelihood of success.

Another challenge involves setting objectives that are unrealistic. For a small or medium-sized enterprise (SME), aiming to completely transform its business through AI could be an unrealistic goal; in these cases, iterative proceess with achievable milestones may be more suitable. Reaching these middle goals not only provides structure but also helps sustain motivation on the project over time.

Finally, if a project lacks relevance to the business, it may always fall to the bottom of the priority list, often resulting in long, unfinished projects. It's crucial to communicate the strategic importance of an AI-based project; in a landscape where this technology is reshaping the economy, delaying AI integration could result in the loss of competitive advantage or even eventual market exit. Indeed, generative AI has the potential to boost labor productivity growth by 0.1 to 0.6 percent annually through 2040, depending on technology adoption rates and the redirection of labor into other activities. When combined with other automation technologies, generative AI could contribute an additional 0.5 to 3.4 percentage points per year to productivity growth. \citep{chui2023economic}

\subsubsection{Human Aspects}
the human element is another key factor in the success of an AI-based project. Internal resistance from staff can pose a significant challenge, which is why involving team members in various phases of the project is essential. When AI implementation affects operational roles by drastically reducing workload, it may lead to concerns about negative potential job impacts. As AI increasingly permeates nearly every professional role, this shift is inevitable. Thus, focusing on upskilling and reshaping roles to redeploy staff into value-added activities is crucial.

Another issue is knowledge: SMEs often lack dedicated resources with expertise in AI, so investing in staff training is vital to support this transition. Finally, company culture plays a central role. Without leadership commitment and a culture that embraces change, advancing an AI implementation project can be very challenging.

\subsubsection{External support}
external support is a crucial factor for the success of AI projects within SMEs. Bridging the gap between small and large enterprises requires the involvement of external actors who can provide targeted guidance and resources. Non-profit organizations, in particular, should play a key role, offering support and guidance, including through specific programs.

Awareness and access to relevant information regarding AI tools and applications are also lacking among many SMEs. Without a clear understanding of how AI can benefit their business processes, decision-makers may be hesitant to invest in these technologies. 

Financially, due to limited resources available to SMEs, initiating AI implementation projects may be challenging. In this context, access to state-provided funds and financing is a necessary aid.

Finally, the advancement of Artificial Intelligence has highlighted the need for a dedicated AI mentor or consultant who can serve as an intermediary between the organization and technology developers. This professional would not only support the organization throughout all stages of implementation but also help guide the project toward long-term objectives, ensuring that investments have a lasting and sustainable impact.

\subsubsection{Technical Enablers}
one of the key challenges in AI implementation is the availability and quality of data, which is often limited or even stored exclusively in paper format. Data quality and quantity are crucial factors in the development and training of an AI system. SMEs often lack the resources to collect or access the vast amounts of data required to train AI systems effectively. Poor or insufficient data may lead to suboptimal models, reducing the potential benefits of AI in automating and enhancing business processes. To address this issue, intermediate solutions could be developed that incorporate an integrated data collection plan. Specifically, Proof of Concepts (PoCs) could be designed to both accumulate data through their usage and to introduce AI within the company on a small scale. These PoCs would be structured to allow scalability for future expansions.

In many cases, AI solutions are difficult to integrate within SMEs due to existing process constraints. Thus, adopting a PoC approach could be combined with the re-engineering of certain processes, ensuring that the integration of the future AI system is feasible. 

Finally, for companies with outdated technological infrastructures, cloud-based AI models provided by external vendors could serve as a practical solution, enabling access to advanced AI capabilities without requiring extensive on-premises infrastructure investments.

\begin{figure}[h!]
    \centering
    \includegraphics[width=0.7\textwidth]{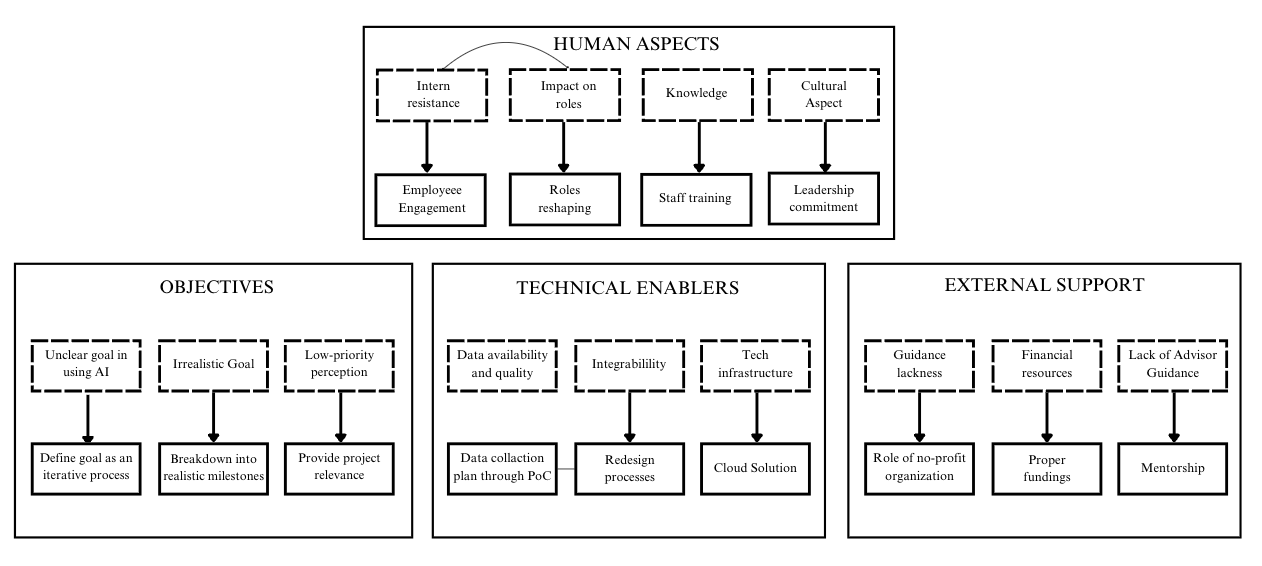}
    \caption{Framework for AI implementation in SMEs} 
    \label{Objectives} 
\end{figure}

\section{Discussion}
The limited sample size implies that the results obtained are not statistically representative and cannot be generalized to the entire target population. 

However, despite the sample limitations, the data collected provide a preliminary indication of the current state of artificial intelligence adoption within this specific group of companies. 

The results obtained should be interpreted as a starting point for other investigations and to stimulate broader discussions on the topic within companies.

For future research, it would be valuable to conduct a deeper analysis of the differences between the Italian business landscape and those of other countries, particularly in the adoption and integration of AI within SMEs. Italy's unique industrial structure and high concentration of family-owned businesses may lead to distinct challenges and opportunities in comparison to other nations. A cross-country study could shed light on best practices and common barriers, potentially guiding more effective AI adoption strategies tailored to local contexts.

Finally, replicating the work conducted in Singapore, where a self-assessment framework was developed to guide AI readiness, could be highly beneficial. Creating a similar self-assessment tool tailored specifically to the needs of Italian SMEs would provide companies with a practical resource to evaluate their AI capabilities and identify gaps. By building this guide on the basis of multiple case studies, the tool could offer insights into typical AI implementation challenges and solutions observed across various sectors. This approach would help companies benchmark themselves against industry standards and develop actionable strategies for AI adoption, thereby fostering a more informed and structured approach to digital transformation in the Italian SME landscape.
\vspace{6pt} 

\bibliographystyle{plainnat} 
\bibliography{ifacconf_latex/root}           

\begin{thebibliography}{10}

\bibitem{bahdanau2014neural}
Dzmitry Bahdanau, Kyunghyun Cho, and Yoshua Bengio.
\newblock Neural machine translation by jointly learning to align and translate.
\newblock {\em arXiv preprint arXiv:1409.0473}, 2014.

\bibitem{chauhan2018convolutional}
Rahul Chauhan, Kamal~Kumar Ghanshala, and RC~Joshi.
\newblock Convolutional neural network (cnn) for image detection and recognition.
\newblock In {\em 2018 first international conference on secure cyber computing and communication (ICSCCC)}, pages 278--282. IEEE, 2018.

\bibitem{chui2023economic}
Michael Chui, Eric Hazan, Roger Roberts, Alex Singla, and Kate Smaje.
\newblock The economic potential of generative ai.
\newblock 2023.

\bibitem{cong2023review}
Shuang Cong and Yang Zhou.
\newblock A review of convolutional neural network architectures and their optimizations.
\newblock {\em Artificial Intelligence Review}, 56(3):1905--1969, 2023.

\bibitem{day2018robotics}
Chia-Peng Day.
\newblock Robotics in industry—their role in intelligent manufacturing.
\newblock {\em Engineering}, 4(4):440--445, 2018.

\bibitem{hansen2021artificial}
Emil~Blixt Hansen and Simon B{\o}gh.
\newblock Artificial intelligence and internet of things in small and medium-sized enterprises: A survey.
\newblock {\em Journal of Manufacturing Systems}, 58:362--372, 2021.

\bibitem{ingalagi2021artificial}
Sanjeev~S Ingalagi, RR~Mutkekar, and PM~Kulkarni.
\newblock Artificial intelligence (ai) adaptation: Analysis of determinants among small to medium-sized enterprises (sme’s).
\newblock In {\em IOP Conference Series: Materials Science and Engineering}, volume 1049, page 012017. IOP Publishing, 2021.

\bibitem{magnani2024}
Roberto Magnani.
\newblock Technological change and professional identity: A study of expectations and concerns in italian knowledge workers, October 31 2024.
\newblock Available at SSRN: \url{https://ssrn.com/abstract=}.

\bibitem{mccarthy2006proposal}
John McCarthy, Marvin~L Minsky, Nathaniel Rochester, and Claude~E Shannon.
\newblock A proposal for the dartmouth summer research project on artificial intelligence, august 31, 1955.
\newblock {\em AI magazine}, 27(4):12--12, 2006.

\bibitem{munro2013sme}
David Munro.
\newblock What is an sme?
\newblock In {\em A guide to SME financing}, pages 7--13. Springer, 2013.

\bibitem{nadkarni2011natural}
Prakash~M Nadkarni, Lucila Ohno-Machado, and Wendy~W Chapman.
\newblock Natural language processing: an introduction.
\newblock {\em Journal of the American Medical Informatics Association}, 18(5):544--551, 2011.

\bibitem{nilsson1984shakey}
Nils~J Nilsson et~al.
\newblock Shakey the robot.
\newblock 1984.

\bibitem{oldemeyer2024investigation}
Leon Oldemeyer, Andreas Jede, and Frank Teuteberg.
\newblock Investigation of artificial intelligence in smes: a systematic review of the state of the art and the main implementation challenges.
\newblock {\em Management Review Quarterly}, pages 1--43, 2024.

\bibitem{isago}
{Personal Data Protection Commission Singapore}.
\newblock Companion to the model ai governance framework – implementation and self-assessment guide for organizations.
\newblock Technical report, Personal Data Protection Commission Singapore (PDPC), 2024.

\bibitem{9908467}
Shavneet Sharma, Gurmeet Singh, Nazrul Islam, and Amandeep Dhir.
\newblock Why do smes adopt artificial intelligence-based chatbots?
\newblock {\em IEEE Transactions on Engineering Management}, 71:1773--1786, 2024.

\bibitem{shum2018eliza}
Heung-Yeung Shum, Xiao-dong He, and Di~Li.
\newblock From eliza to xiaoice: challenges and opportunities with social chatbots.
\newblock {\em Frontiers of Information Technology \& Electronic Engineering}, 19:10--26, 2018.

\bibitem{vaswani2017attention}
Ashish Vaswani, Noam Shazeer, Niki Parmar, Jakob Uszkoreit, Llion Jones, Aidan~N. Gomez, Lukasz Kaiser, and Illia Polosukhin.
\newblock Attention is all you need, 2017.

\end{thebibliography}

\end{document}